\documentclass{article}
\usepackage{ijcai22}

\usepackage{times}
\usepackage{soul}
\usepackage{url}
\usepackage[hidelinks]{hyperref}
\hypersetup{hidelinks}
\usepackage[utf8]{inputenc}
\usepackage[small]{caption}
\usepackage{graphicx}
\usepackage{amsmath}
\usepackage{amsthm}
\usepackage{booktabs}
\usepackage{algorithm}
\usepackage{algorithmic}
\usepackage{amsbsy}
\usepackage{booktabs}
\usepackage[table,xcdraw]{xcolor}
\usepackage{verbatim}

\makeatletter
\newcommand{\pushright}[1]{\ifmeasuring@#1\else\omit\hfill$\displaystyle#1$\fi\ignorespaces}
\makeatother

\title{Safe Reinforcement Learning via Probabilistic Logic Shields}

\author{
Wen-Chi Yang$^1$
\and
Giuseppe Marra$^1$\and
Gavin Rens\And
Luc De Raedt$^{1,2}$
\affiliations
$^1$Leuven AI, KU Leuven, Belgium\\
$^2$Centre for Applied Autonomous Sensor Systems, \"Orebro University, Sweden.
\emails
\{wenchi.yang, giuseppe.marra, luc.deradet\}@kuleuven.be,
gavin.rens@gmail.com
}

\usepackage{amssymb} 
\usepackage[capitalize]{cleveref}

\newtheorem{theorem}{Theorem}[section]
\newtheorem{definition}[theorem]{Definition}
\newtheorem{proposition}{Proposition}

\usepackage[inline]{enumitem}
\usepackage{amsfonts} 
\usepackage{bbm}
\usepackage{xcolor}
\usepackage{fancyvrb} 
\usepackage{xspace}
\usepackage{svg}
\usepackage{subcaption}
\usepackage{multirow}
\setlist{nosep}

\newcommand{\problog}{ProbLog\xspace}

\newcommand{\PPO}{\texttt{PPO}\xspace}
\newcommand{\PLPG}{\texttt{PLPG}\xspace}
\newcommand{\VSRL}{\texttt{VSRL}\xspace}

\begin{document}

\maketitle

\begin{abstract}
Safe Reinforcement learning (Safe RL) aims at learning optimal policies while staying safe. A popular solution to Safe RL is shielding, which uses a logical safety specification to prevent an RL agent from taking unsafe actions. However, traditional shielding techniques are difficult to integrate with continuous, end-to-end deep RL methods. To this end, we introduce Probabilistic Logic Policy Gradient (PLPG). PLPG is a model-based Safe RL technique that uses probabilistic logic programming to model logical safety constraints as differentiable functions. Therefore, PLPG can be seamlessly applied to any policy gradient algorithm while still providing the same convergence guarantees. In our experiments, we show that PLPG learns safer and more rewarding policies compared to other state-of-the-art shielding techniques.
\end{abstract}

\section{Introduction}

Shielding is a popular Safe Reinforcement Learning (Safe RL) technique that aims at finding an optimal policy while staying safe~\cite{Shielding2020}. 
To do so, it relies on a \textit{shield}, a logical component that monitors the agent's actions and rejects those that violate the given safety constraint.
These {\em rejection-based} shields are typically based on formal verification, offering strong safety guarantees compared to other safe exploration techniques~\cite{Garcia2015}.
While early shielding techniques operate completely on symbolic state spaces~\cite{Shielding2020,DetShield2017,Bastani2018}, more recent ones have incorporated a neural policy learner to handle continuous state spaces~\cite{HardNeuralShielding2020,Anderson2020,Harris2020}. In this paper, we will also focus {\em  on integrating shielding with neural policy learners.}

\begin{figure*}[t]
    \centering
    \includegraphics[width=\textwidth]{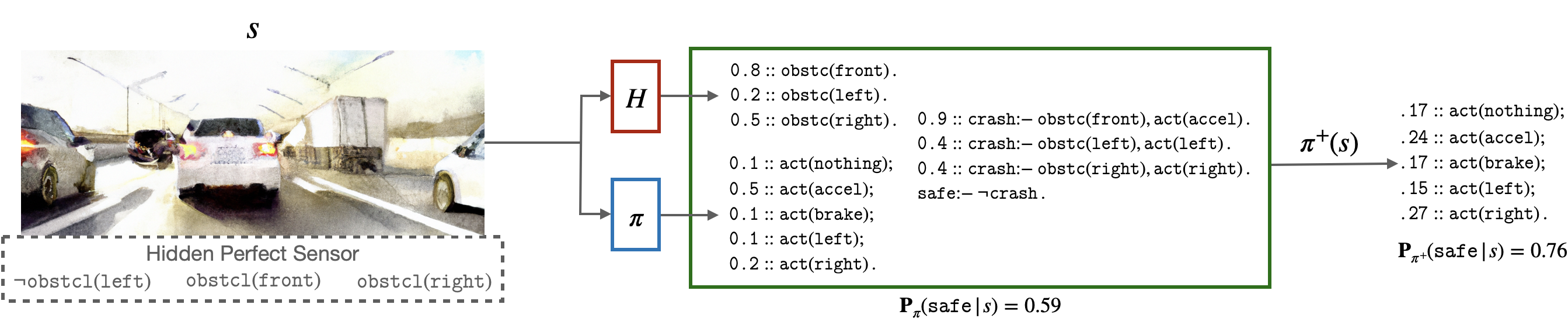}
    \caption{
    A motivation example of Probabilistic Logic Shields. 
    We encode the interaction between the base policy $\pi$, the state-abstraction $H$ and the safety specification using a ProbLog program $\mathcal{T}$. This provides a uniform language to express many aspects of the shielding process. 
    The shielded policy $\pi^+(s)$ decreases the probability of unsafe actions, e.g. acceleration and increases the likelihood of being safe. 
    }
    \label{fig: motivating_example}
\end{figure*}

In current shielding approaches, the shields are deterministic, that is, an action is either safe or unsafe in a particular state. 
This is an unrealistic assumption as the world is inherently uncertain and safety is a matter of degree and risk rather than something absolute.
For example, \cref{fig: motivating_example} demonstrates a scenario in which a car must detect obstacles from visual input and the sensor readings are noisy. 
The uncertainty arising from noisy sensors cannot be directly exploited by such rejection-based shields. 
In fact, it is often assumed that agents have perfect sensors~\cite{Giacobbe2021,HardNeuralShielding2020}, which is unrealistic. 
By working with probabilistic rather than deterministic shields, \textit{we will be able to cope with such uncertainties and risks}.

Moreover, even when given perfect safety information in all states, rejection-based shielding may fail to learn an optimal policy~\cite{Ray2019,HardNeuralShielding2020,Anderson2020}. 
This is due to the way the shield and the agent interact as the learning agent is not aware of the rejected actions and continues to update its policy as if all safe actions were sampled directly from its safety-agnostic policy instead of through the shield. 
By eliminating the mismatch between the shield and the policy, \textit{we will be able to guarantee  convergence towards an optimal policy if it exists}.

We introduce \textit{probabilistic shields} as an alternative to the deterministic rejection-based shields. 
Essentially, probabilistic shields take the original policy and noisy sensor readings to produce a safer policy, as demonstrated in \cref{fig: highlevel} (right).
By explicitly connecting action safety to probabilistic semantics, probabilistic shields provides a realistic and principled way to balance return and safety. 
This also allows for shielding to be applied at the level of the policy instead of at the level of individual actions, which is typically done in the literature \cite{HardNeuralShielding2020,Shielding2020}.

We propose the concept of \textit{Probabilistic Logic Shields (PLS)} and its implementation in probabilistic logic programs.
Probabilistic logic shields are probabilistic shields that models safety through the use of logic.
The safety specification is expressed as background knowledge and its interaction with the learning agent and the noisy sensors is encoded in   a probabilistic logic program, which is an elegant and effective way of defining a shield. 
Furthermore, PLS can be automatically compiled into a differentiable structure, allowing for the optimization of a single loss function through the shield, enforcing safety directly in the policy. 
Probabilistic logic shields have several benefits:  
\begin{itemize}
\item \textit{A Realistic Safety Function}. PLS models a more realistic, probabilistic evaluation of safety instead of a deterministic one, it allows to balance safety and reward, and in this way control the risk. 
\item \textit{A Simpler Model.}
PLS do not require for knowledge of the underlying MDP, but instead use a simpler safety model that only represents internal safety-related properties. 
This is less demanding than requiring the full MDP be known
required by many model-based approaches \cite{Shielding2020,HardNeuralShielding2020,Carr2022}. 
\item \textit{End-to-end Deep RL.}
By being differentiable, PLS allows for seamless application of probabilistic logic shields to any model-free reinforcement learning agent such as PPO~\cite{Schulman2018}, TRPO~\cite{Schulman2015}, A2C~\cite{Mnih2016}, etc. 
\item \textit{Convergence.} Using PLS in deep RL comes with convergence guarantees unlike the use of rejection-based shields.
\end{itemize}

\section{Preliminaries}

\paragraph{Probabilistic Logic Programming}
\label{sec: problog}

We will introduce probabilistic logic programming (PLP) using the syntax of the ProbLog system \cite{problogpaper}.
An  \textit{atom} is a predicate symbol followed by a tuple of logical variables and/or constants.
A \textit{literal} is an atom or its negation. 
A \problog theory (or program) $\mathcal{T}$ consists of a finite set of probabilistic facts $\mathcal{F}$ and a finite set of clauses $\mathcal{BK}$. 
A \textit{probabilistic fact} is an expression of the form $\mathtt{p_i::f_i}$ where $\mathtt{p_i}$ denotes the probability of the fact $\mathtt{f_i}$ being true. It is assumed that the probabilistic facts are independent of one other (akin to the nodes with parents in Bayesian networks) and the dependencies are specified by clauses (or rules).   For instance, $\mathtt{0.8::obstc(front).}$ states
 that the probability of having an obstacle in front is 0.8.
A \textit{clause} is a universally quantified expression of the form $\mathtt{h :\!\!- b_1, ...,b_n}$ where $\mathtt{h}$ is a literal and $\mathtt{b_1, ...,b_n}$ is a conjunction of literals, stating that $\mathtt{h}$ is true if all $\mathtt{b_i}$ are true. 
    The clause defining $\mathtt{safe}$ in Fig. \ref{fig: motivating_example} states that it is safe when there is no crash.
Each truth value assignment of all probabilistic facts $\mathcal{F}$, denoted by $\mathcal{F}_k$, induces a possible world 
$w_k$ where all ground facts in $w_k$ are \textit{true} and all that are not in $w_k$ are \textit{false}. 
    Formally, the probability of a possible world $w_k$ is defined as follows.
    \begin{align}\label{eq: world probability}
    	P(w_k) =
    	\prod_{\mathtt{f_i}\in w_k} \mathtt{p_i} 
    	\prod_{\mathtt{f_i}\not\in w_k} \mathtt{(1-p_i)}.
    \end{align}
\problog allows for \textit{annotated disjunctions} (ADs). An AD is a clause with multiple heads $\mathtt{h_i}$ that are mutually exclusive to one another, meaning that exactly one head is true when the body is true, and the choice of the head is governed by a probability distribution. 
An AD has the form of $\mathtt{p_1::h_1; \cdots;p_m::h_m}$ where 
each $\mathtt{h_i}$ is the head of a clause and
$\sum^m_{i=1} \mathtt{p_i}\leq 1$~\cite{problogpaper,problog22015}.
For instance, 
$\{\mathtt{0.1::act(nothing);}$ 
$\mathtt{0.5::act(accel);}$ 
$\mathtt{0.1::act(brake);}$ 
$\mathtt{0.1::act(left);}$ 
$\mathtt{0.2::act(right)}\}$ is an AD (with no conditions), stating the probability with which each action will be taken. 
The \textit{success probability} of an atom $q$ given a theory $\mathcal{T}$ is the sum of the probabilities of the possible worlds that entail $q$.
Formally, it is defined as $P(q) := \sum_{w_k \models q} P(w_k).$
Given a set of atoms $E$ as evidence, the \textit{conditional probability} of a query $q$ is $\mathbf{P}(q|E) = \frac{P(q,E)}{P(E)}$. 
For instance, in \cref{fig: motivating_example}, the probability of being safe in the next state given that the agent accelerates is 
$
\mathbf{P}(\mathtt{safe|act(accel)})= \frac{0.14}{0.5} = 0.28
$.

\label{sec: shielding}

\paragraph{MDP} A Markov decision process (MDP) is a tuple $M=\langle S, A, T, R ,\gamma\rangle$ where 
$S$ and $A$ are state and action spaces, respectively. 
$T(s,a,s') = P(s'|s,a)$ defines the probability of being in $s'$ after executing $a$ in $s$.
$R(s,a,s')$ defines the immediate reward of executing $a$ in $s$ resulting in $s'$. 
$\gamma\in [0,1]$ is the discount factor. A policy $\pi: S \times A \to [0,1]$ defines the probability of taking an action in a state.
We use $\pi(a |s)$ to denote the probability of action $a$ in state $s$ under policy $\pi$ and  $\pi(s)$ for the probability distribution over all the actions in state $s$.

\paragraph{Shielding} A shield is a reactive system that guarantees safety of the learning agent by preventing it from selecting any unsafe action at run time~\cite{Shielding2020}. 
Usually, a shield is provided as a Markov model and a temporal logical formula, jointly defining a safety specification.
The shield constrains the agent's exploration by only allowing it to take actions that satisfy the given safety specification. 
For example, when there is a car driving in front and the agent proposes to \textit{accelerate}, a standard shield (\cref{fig: highlevel}, left) will reject the action as the agent may crash.

\paragraph{Policy Gradients} 
The objective of a RL agent is to find a policy $\pi_\theta$ (parameterized by $\theta$) that maximizes the total expected return along the trajectory, formally,
$$
J(\theta) = \mathbb{E}_{\pi_\theta}[\sum_{t=0}^{\infty} \gamma^t r_t]$$
$J(\theta)$ is assumed to be finite for all policies. 
An important class of RL algorithms, particularly relevant in the setting of shielding, is based on \textit{policy gradients}, which maximize $J(\theta)$ by repeatedly estimating the gradient $\nabla_\theta J(\theta)$. The general form of policy gradient methods is:
\begin{align}
\label{eq: policy gradient}
    \nabla_\theta J(\theta) = \mathbb{E}_{\pi_\theta}[\sum_{t=0}^{\infty} \Psi_t \nabla_\theta \log \pi_\theta(s_t,a_t)]
\end{align}
where $\Psi_t$ is an empirical expectation of the return~\cite{Schulman2018}. The expected value is usually computed using Monte Carlo methods, requiring efficient sampling from $\pi_\theta$.

\section{Probabilistic Logic Shields}

\subsection{Probabilistic Shielding}
To reason about safety, we will assume that we have a safety-aware probabilistic model $\mathbf{P}(\mathtt{safe} | a, s)$, indicating the probability that an action $a$ is safe to execute in a state $s$. We use bold $\mathbf{P}$ to distinguish it from the underlying probability distributions $P(\cdot)$ of the MDP. Our safety-aware probabilistic model does not require that the underlying  MDP is known. However, it needs to represent internally safe relevant-dynamics as the safety specification. 
Therefore, the safety-aware model $\mathbf{P}$ is not a full representation of the MDP, it is  limited to safety-related properties.
 
Notice that for traditional logical shields actions are either safe or unsafe, which
implies that $\mathbf{P}(\mathtt{safe} | a, s) = 1$ or $0$.
We are interested in 
the safety of policies, that is, in
\begin{align} \mathbf{P}_\pi(\mathtt{safe} | s) = \sum_{a\in A} \mathbf{P} (\mathtt{safe}|s,a)\pi(a|s).\label{eq: policy safety}
\end{align}

By marginalizing out actions according to their probability under $\pi$, the probability $\mathbf{P}_\pi$ measures how likely it is to take a safe action in $s$ according to $\pi$.
\begin{definition}\textbf{Probabilistic Shielding }
\label{def: differentiable shield}
Given a \textit{base policy} $\pi$ and a  probabilistic safety model $\mathbf{P}(\mathtt{safe} | s, a)$, the \textit{shielded policy} is
\begin{align}
\label{eq: shielded policyy}
\pi^+(a|s)
    &= \mathbf{P}_\pi(a|s, \mathtt{safe} ) =\frac{\mathbf{P}(\mathtt{safe} | s, a)}{\mathbf{P}_{\pi}(\mathtt{safe} |s)} \pi(a|s)
\end{align}
\end{definition}

Intuitively, a shielded policy is a re-normalization of the base policy that increases (resp. decreases) the probabilities of the actions that are safer (resp. less safe) than average. 

\begin{proposition}
    A shielded policy is always safer than its base policy in all states, i.e.\ $P_{\pi^+}(\mathtt{safe}|s) \geq P_\pi(\mathtt{safe}|s)$ for all $s$ and $\pi$. (A proof is in \cref{app: proof})
\end{proposition}

\subsection{Shielding with Probabilistic Logics}
In this paper, we focus on probabilistic shields implemented through \textit{probabilistic logic programs}. The ProbLog program defining  probabilistic safety consists of three parts.

First, an annotated disjunction $\mathbf{\Pi}_s$, which   represents the policy $\pi(a|s)$ in the logic program. For example, in \cref{fig: motivating_example}, the policy $\pi(a | s)$ is represented as the annotated disjunction $\mathbf{\Pi}_s $ 
$=\{\mathtt{0.1::act(nothing);}$ 
$\mathtt{0.5::act(accel);}$ 
$\mathtt{0.1::act(brake);}$ 
$\mathtt{0.1::act(left);}$ 
$\mathtt{0.2::act(right)}\}$.

The second component of the program is a set of probabilistic facts $\mathbf{H}_s$ representing an  abstraction of the current state.
Such abstraction should contain information needed to reason about the safety of actions in that state. Therefore, it should 
not be a fair representation of the entire state $s$. 
For example, in \cref{fig: motivating_example}, the 
abstraction is represented as the set of probabilistic facts: $\mathbf{H}_s = \{$ $\mathtt{0.8::obstc(front).}$ $\mathtt{ 0.2::obstc(left).}$ $\mathtt{  0.5::obstc(right)}.\}$

The third component of the program is a safety specification $\mathcal{BK}$, which is a set of  clauses and ADs representing  knowledge about safety. 
In our example, the predicates $\mathtt{crash}$ and $\mathtt{safe}$ are defined using clauses. The first clause for $\mathtt{crash}$ states that the probability of having a crash is 0.9 if the agent accelerates when there is an obstacle in front of it. The clause for $\mathtt{safe}$ states that it is safe if no crash occurs.

Therefore, we obtain a ProbLog program 
$\mathcal{T}(s)= \mathcal{BK} \cup \mathbf{H}_s \cup \boldsymbol{\Pi}_s$, inducing a probabilistic measure $\textbf{P}_\mathcal{T}$. By querying this program, we can reason about safety of policies and actions.
More specifically, we can use $\mathcal{T}(s)$ to obtain:
\begin{itemize}
\item
action safety in $s$: 
$\textbf{P}(\mathtt{safe}|s,a) = \mathbf{P}_\mathcal{T}(\mathtt{safe} | a)$
\item
policy safety in $s$: $\textbf{P}_\pi(\mathtt{safe}|s) = \mathbf{P}_\mathcal{T}(\mathtt{safe})$
\item
the shielded policy in $s$: $\textbf{P}_\pi(a|s,\mathtt{safe}) =
\mathbf{P}_\mathcal{T}(a|\mathtt{safe})$
\end{itemize}
Notice that these three distributions can be obtained by querying the same identical program ProbLog program $\mathcal{T}$.

It is important to realize that although we are using 
the probabilistic logic programming language ProbLog to reason
about safety, we could also have used alternative representations
such as Bayesian networks, or other StarAI models \cite{DeRaedt16}. 
The ProbLog representation is however convenient because
it allows to easily model planning domains, it is Turing equivalent and
it is differentiable. 
At the same time, it is important to note that the safety model in ProbLog
is an abstraction, using possibly a different representation than
that of the underlying MDP. 

\paragraph{Perception Through Neural Predicates}

The probabilities of both $\mathbf{\Pi}_s$ and $\mathbf{H}_s$ depend on the current state $s$. 
We compute these probabilities using two neural networks $\pi$ and $H$
operating on real valued inputs (i.e. images or sensors). We feed then the probabilities to the ProbLog program. As we depicted in Fig. \ref{fig: motivating_example}, both these functions take a state representation $s$ as input (e.g. an image) and they output the probabilities the facts representing, respectively, the actions and the safe-relevant abstraction of the state.
This feature is closely related to the notion of \textit{neural predicate}, which can — for the purposes of this paper — be regarded as encapsulated neural networks inside logic. 
Since ProbLog programs are differentiable with respect to the probabilities in $\mathbf{\Pi}_s$ and $\mathbf{H}_s$, the gradients can be seamlessly backpropagated to network parameters during learning.

\section{Probabilistic Logic Policy Gradient}

\begin{figure}[t]
    \centering
    \includegraphics[width=\columnwidth]{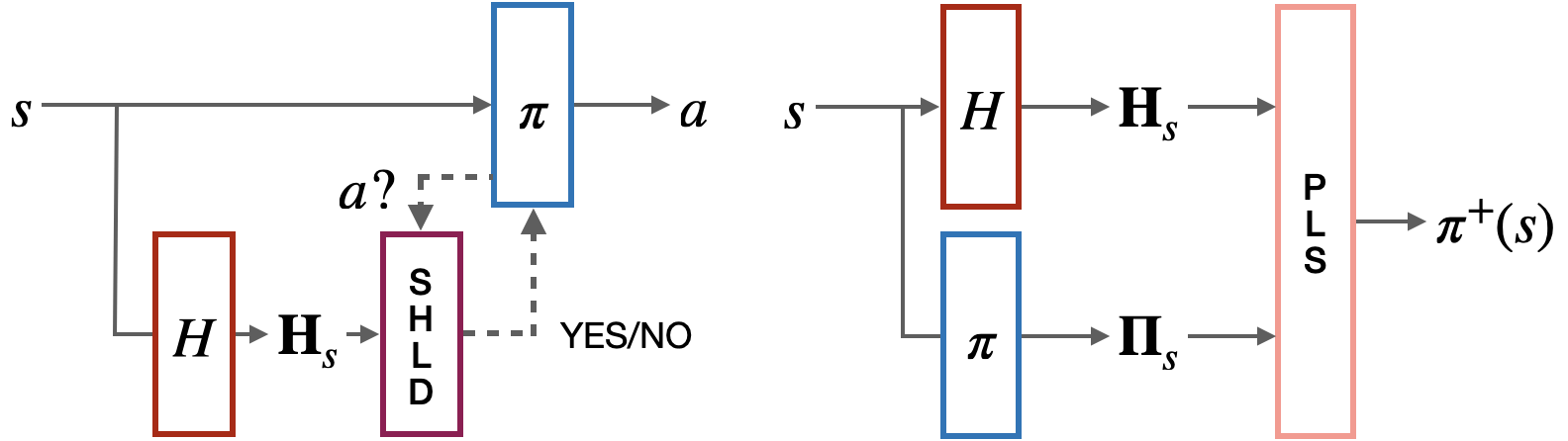}
    \caption{ 
    A comparison between a traditional rejection-based shield (\texttt{SHLD}) and a Probabilistic Logic Shield (\texttt{PLS}). Both shields take a policy $\pi$ and a set of noisy sensor readings $\mathbf{H_s}$ to compute a safe action.
    \textbf{Left}: The policy must keep sampling actions until a safe action is accepted by the rejection-based shield. This requires an assumption that an action is either completely safe or unsafe.
    \textbf{Right}: We replace \texttt{SHLD} with \texttt{PLS} that proposes a safer policy $\pi^+$ on the policy level without imposing the assumption.
    }
    \label{fig: highlevel}
\end{figure}

In this section, we demonstrate how to use probabilistic logic shields with a deep reinforcement learning method. 
We will focus on real-vector states (such as images), but the safety constraint is specified symbolically and logically. 
Under this setting, our goal is to find an optimal policy in the safe policy space $\Pi_\mathtt{safe}=\{\pi\in\Pi | \mathbf{P}_\pi(\mathtt{safe}|s) =1, \forall s\in S\} $. 
Notice that it is possible that no optimal policies exist in the safe policy space. In this case, our aim will be to find a policy that it as safe as possible. 

We propose a new safe policy gradient technique, which we call \textit{Probabilistic Logic Policy Gradient (PLPG)}. PLPG applies probabilistic logic shields and is guaranteed to converge to a safe and optimal policy if it exists.

\subsection{PLPG for Probabilistic Shielding}
\label{sec: prob safety}

To integrate PLS with policy gradient, we simply replace the base policy in \cref{eq: policy gradient} with the shielded policy obtained through \cref{eq: shielded policyy}. We call this a \textit{shielded policy gradient}.
\begin{align}\label{eq: shielded policy gradient}
    \mathbb{E}_{\pi^+_\theta}[&\sum_{t=0}^{\infty} \Psi_t \nabla_\theta \log \pi^+_\theta(a_t|s_t)]
\end{align}
This requires the shield to be differentiable and cannot be done using rejection-based shields. 
The gradient encourages policies $\pi^+$ that are more rewarding, i.e. that has a large $\Psi$, in the same way of a standard policy gradient. It does so by assuming that unsafe actions have been filtered by the shield. However, when the safety definition is uncertain, unsafe actions may still be taken and such a gradient may still end up encouraging unsafe but rewarding policies.

For this reason, we introduce a safety loss to penalize unsafe policies.
Intuitively, a safe policy should have a small safety loss, and the loss of a completely safe policy should be zero. 
The loss can be expressed by interpreting the policy safety as \textit{the probability that $\pi^+$ satisfies the safety constraint} (using
the semantic loss of \cite{SemanticLoss2018}); more formally, $L^s := -\log \mathbf{P}_{\pi^+}(\mathtt{safe}|s)$. 
Then, the corresponding \textit{safety gradient} is as follows. 
\begin{align}\label{eq: safety gradient}
     -\mathbb{E}_{\pi^+_\theta}[\nabla_\theta\log \mathbf{P}_{\pi^+}(\mathtt{safe}|s)]
\end{align}
Notice that the safety gradient is not a shielding mechanism but a regret due to its loss-based nature. 
By combining the shielded policy gradient and the safety gradient, we obtain a new Safe RL technique. 
\begin{definition}\textbf{(PLPG)} The probabilistic logic policy gradient $\nabla_\theta J(\theta)$ is 
{\small
\begin{align}
\label{eq: PLPG}
     \mathbb{E}_{\pi^+_\theta}[&\sum_{t=0}^{\infty} \Psi_t \nabla_\theta \log \pi^+_\theta(a_t|s_t) 
    - \alpha\nabla_\theta\log \mathbf{P}_{\pi^+_\theta}(\mathtt{safe}|s_t)]
\end{align}
}%
where $\alpha$ is the safety coefficient that indicates the weight of the safety gradient. 
\end{definition}

We introduce the safety coefficient $\alpha$, a hyperparameter that controls the combination of the two gradients. 

Both gradients in PLPG are essential. 
The shielded policy gradient forbids the agent from immediate danger and the safety gradient penalizes unsafe behavior. 
The interaction between shielded and loss-based gradients is still a very new topic, nonetheless, recent advances in neuro-symbolic learning have shown that both are important \cite{Ahmed2022}. 
Our experiments shows that having only one but not both is practically insufficient.

\subsection{Probabilistic vs Rejection-based Shielding}
The problem of finding a policy in the safe policy space cannot be solved by rejection-based shielding (cf. \cref{fig: highlevel}, left) without strong assumptions, such as that a state-action pair is either completely safe or unsafe. 
It is in fact often assumed that a set of safe state-action pairs $SA_{\mathtt{safe}}$ is given. 
To implement a rejection-based shield, the agent must keep sampling from the base policy until an action $a$ is accepted. 
This approach implicitly conditions through very inefficient rejection sampling schemes as below, which has an unclear link to probabilistic semantics~\cite{robert1999monte}. 
\begin{align}
    \pi^+(a|s) = 
    \frac{\mathbbm{1}_{[(s,a)\in SA_{\mathtt{safe}}]}\pi(a|s)}{\sum\limits_{a'\in A}\mathbbm{1}_{[(s,a')\in SA_{\mathtt{safe}}]}\pi(a'|s)}
    \label{eq: re-noremalizetion}
\end{align}

\subsection*{Convergence Under Perfect Safety Information}

It is common for existing shielding approaches to integrate policy gradients with a rejection-based shield, e.g. \cref{eq: re-noremalizetion}, resulting in the following policy gradient.  
\begin{align}\label{eq: classical gradient}
    \nabla_\theta J(\theta) = \mathbb{E}_{\pi^+_\theta}[\sum_{t=0}^{\infty} \Psi_t \nabla_\theta \log \pi_\theta(s_t,a_t)]
\end{align}

It is a known problem that this approach may result in sub-optimal policies, even though the agent has perfect safety information~\cite{Kalweit2020,Ray2019,HardNeuralShielding2020,Anderson2020}.
This is due to a policy mismatch between $\pi^+$ and $\pi$ in \cref{eq: classical gradient}.
More specifically, \cref{eq: classical gradient} is
an \textit{off-policy} algorithm, meaning that the policy used to explore (i.e.\ $\pi^+$) is different from the policy that is updated (i.e.\ $\pi$)\footnote{A well-known off-policy technique is Q-learning as the behavior policy is $\epsilon$-greedy compared to the target policy. }. 
For any off-policy policy gradient method to converge to an optimal policy (even in the tabular case), the behavior policy (to explore) and the target policy (to be updated) must appropriately visit the same state-action space, i.e., if $\pi(a|s)>0$ then $\pi^+(a|s)>0$~\cite{RLbook}. This requirement is violated by \cref{eq: classical gradient}. 

PLPG, on the other hand, reduces to \cref{eq: shielded policy gradient} when given perfect sensor information of the state.
Since it has the same form as the standard policy gradient (Eq.~\ref{eq: classical gradient}), PLPG
is guaranteed to converge to an optimal policy according to the Policy Gradient Theorem~\cite{Sutton2000}. 
\begin{proposition}\label{prop: convergence}
    PLPG, i.e. \cref{eq: PLPG}, converges to an optimal policy given perfect safety information in all states.
\end{proposition}

It should be noted that the base policy learnt by PLPG is unlikely to be equivalent to the policy that would have been learnt without a shield, all other things being equal. That is, in general, $\pi_\theta$ in~\cref{eq: classical gradient} will not tend to be the same as $\pi_\theta$ in ~\cref{eq: PLPG} as learning continues. 
This is because the parameters $\theta$ in~\cref{eq: PLPG} depend on $\pi^+_\theta$ instead of $\pi_\theta$, and $\pi_\theta$ is learnt in a way to make $\pi^+$ optimal \textit{given} safety imposed by the shield.

\subsection*{Learning From Unsafe Actions}
PLPG has a significant advantage over a rejection-based shield by learning not only from the actions accepted by the shield but also from the ones that would have been rejected. 
This is a result of the use of ProbLog programs that computes the safety of \textit{all available actions} in the current state,
allowing us to update the base policy in terms of safety without having to actually execute the other actions.

\subsection*{Leveraging Probabilistic Logic Programming}
Probabilistic logic programs can, just like Bayesian networks, be compiled into circuits using knowledge compilation~\cite{Darwiche}.
Although probabilistic inference is hard (\#P-complete), 
once the circuit is obtained, inference is linear in the size of the circuit~\cite{problogpaper,problog22015}. 
Furthermore, the circuits can be used to compute gradients and the safety model only needs to be compiled once. 
We show a compiled program in \cref{app: circuit}. 
Our PLP language is still restricted to discrete actions,
although it should be possible to model continuous actions using extensions thereof in future work~\cite{Nitti2016}.

\section{Experiments}
\label{sec: experiments}

Now, we present an empirical evaluation of PLPG, when compared to other baselines in terms of  return and safety.

\paragraph{Experimental Setup} 

Experiments are run in three environments.
\begin{enumerate*}[label=(\arabic*)]
    \item \textit{Stars}, the agent must collect as many stars as possible without going into a stationary fire;
    \item \textit{Pacman}, the agent must collect stars without getting caught by the fire ghosts that can move around, and  
    \item \textit{Car Racing}, the agent must go around the track without driving into the grass area.
\end{enumerate*} 
We use two configurations for each domain. 
A detailed description is in \cref{app: experiments}.

\begin{figure}[ht]
    \centering
    \raisebox{-\height}{\includegraphics[width=0.3\columnwidth]{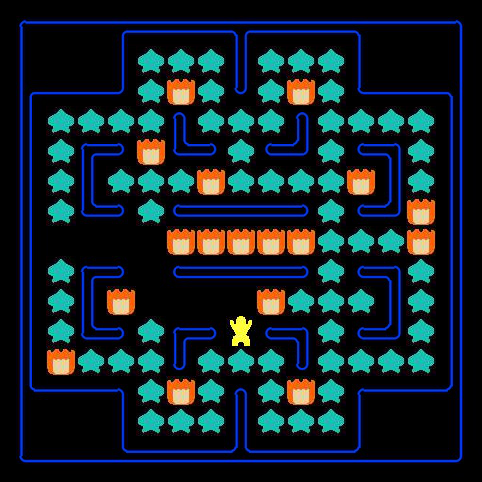}}
    \raisebox{-\height}{\includegraphics[width=0.3\columnwidth]{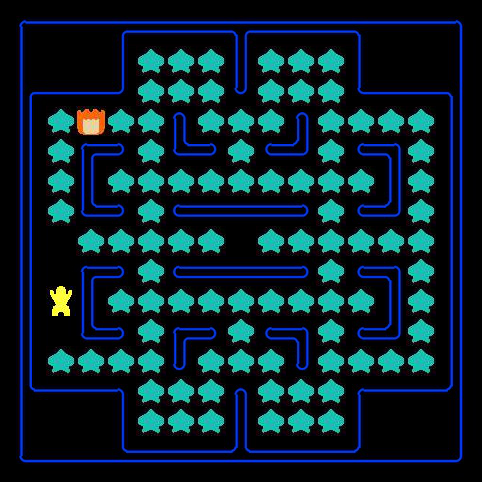}}
    \raisebox{-\height}{\includegraphics[width=0.3\columnwidth]{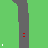}}%
    \caption{The domains of Stars, Pacman and Car Racing. The difference between Stars and Pacman is that fires in Stars are stable and the ones in Pacman move around.  
    }
    \label{fig: smplify image}
\end{figure}

We compare \PLPG to three RL baselines.

\begin{itemize}
\item
\PPO, a standard safety-agnostic agent that starts with a random initial policy~\cite{Schulman2017}.
\item 
\VSRL, an agent augmented with a deterministic rejection-based shield~\cite{HardNeuralShielding2020}. Its structure is shown in \cref{fig: highlevel} (left). 
\item 
$\epsilon$-\VSRL a new 
 risk-taking variant of \VSRL that has a small $\epsilon$ probability of accepting any action, akin to $\epsilon$-greedy~\cite{Fernandez2006}
As $\epsilon$-\VSRL simulates an artificial noise/distrust in the sensors, we expect that it improves \VSRL's ability to cope with noise.
\end{itemize}

In order to have a fair comparison, \VSRL, $\epsilon$-\VSRL and \PLPG agents have identical safety sensors and safety constraints, and \PPO does not have any safety sensors. 
We do not compare to shielding approaches that require a full model of the environment since they are not applicable here. 
We train all agents in all domains using 600k learning steps and all experiments are repeated using five different seeds. 

In our experiments, we answer the following questions :
\begin{itemize}
\item[Q1]  \textit{Does \PLPG produce safer and more rewarding policies than its competitors?} 
\item[Q2] \textit{What is the effect of the hyper-parameters $\alpha$ and $\epsilon$?}
\item[Q3] \textit{Does the shielded policy gradient or the safety gradient have more impact on safety in \PLPG?} 
\item[Q4] \textit{What are the performance and the computational cost of a multi-step safety look-ahead PLS?}
\end{itemize}


\paragraph{Metrics} We compare all agents in terms of two metrics: 
\begin{enumerate*}[label=(\arabic*)]
    \item \textit{average normalized return}, i.e.\ the standard per-episode normalized return averaged over the last 100 episodes;
    \item \textit{cumulative normalized violation}, i.e.\ the accumulated number of constraint violations from the start of learning. 
\end{enumerate*}
The absolute numbers are listed in \cref{app: experiment details}.

\paragraph{Probabilistic Safety via Noisy Sensor Values}
We consider a set of noisy sensors around the agent for safety purposes. For instance, the four fire sensor readings in \cref{fig: smplify image} (left) might 
$\{\mathtt{0.6::fire( 0, 1).}$ $\mathtt{0.1::fire( 0,-1).}$ $\mathtt{0.1::fire(-1, 0).}$ $\mathtt{0.4::fire( 1, 0).}\}$
These sensors provide to the shield only local information of the learning agent. 
\PLPG agents are able to directly use the noisy sensor values. 
However, as \VSRL and $\epsilon$-\VSRL agents require 0/1 sensor values to apply formal verification, noisy sensor readings must be discretized to $\{0, 1\}$. 
For example, the above fire sensor readings become
$\{\mathtt{1::fire( 0, 1).}$ $\mathtt{0::fire( 0,-1).}$ $\mathtt{0::fire(-1, 0).}$ $\mathtt{0::fire( 1, 0).}\}$
For all domains, the noisy sensors are standard, pre-trained neural approximators of an accuracy higher than $99\%$. The training details are presented in \cref{app: experiments}. Even with such accurate sensors, we will show that discretizing the sensor values, as in \VSRL, is harmful for safety. Details of the pre-training procedure are in \cref{app: experiments}.

\subsection*{Q1: Lower Violation and Higher Return} 

We evaluate how much safer and more rewarding \PLPG is compared to baselines by measuring the cumulative safety violation and the average episodic return during the learning process. 
Before doing so, we must select appropriate hyperparameters for \PLPG and \VSRL agents. 
Since our goal is to find an optimal policy in the safe policy space, we select the $\alpha\in\{0, 0.1, 0.5, 1, 5\}$ and $\epsilon\in\{0, 0.005, 0.01, 0.05, 0.1, 0.2, 0.5, 1.0\}$ values that result in the lowest violation in each domain. 
The hyperparameter choices are listed in \cref{app: experiment details}. These values are fixed for the rest of the experiments (except for Q2 where we explicitly analyze their effects). 
We show that \PLPG achieves the lowest violation while having a comparable return to other agents.

\begin{figure}[t]
    \centering
    \includegraphics[width=\columnwidth]{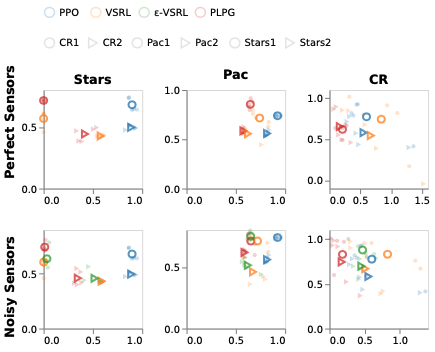}
    \caption{Trade-off between \textbf{Violation (x-axis) and Return (y-axis)}. Each small data point is an agent's policy and the large data points are the average of five seeds. An ideal policy should lie in the upper-left corner. }\label{fig: return-violation}
\end{figure}

The results are plotted in \cref{fig: return-violation} where each data point is a policy trained for 600k steps. 
\cref{fig: return-violation} clearly shows that \PLPG is the safest in most tested domains.
When augmented with perfect sensors, \PLPG's violation is lower than \PPO by 50.2\% and lower than \VSRL by 25.7\%. 
When augmented with noisy sensors, \PLPG's violation is lower than \PPO by 51.3\%, lower than \VSRL by 24.5\% and lower than $\epsilon$-\VSRL by 13.5\%.
\cref{fig: return-violation} also illustrates that \PLPG achieves a comparable (or slightly higher) return while having the lowest violation. 
When augmented with perfect sensors, \PLPG's return is higher than \PPO by 0.5\% and higher than \VSRL by 4.8\%. 
When augmented with noisy sensors, \PLPG's return is higher than \PPO by 4.5\%, higher than \VSRL by 6.5\% and higher than $\epsilon$-\VSRL by 6.7\%.

Car Racing has complex and continuous action effects compared to the other domains. 
In CR, safety sensor readings alone are not sufficient for the car to avoid driving into the grass as they do not capture the inertia of the car, which involves the velocity and the underlying physical mechanism such as friction. 
In domains where safety sensors are not sufficient, the agent must be able to act a little unsafe to learn, as \PPO, $\epsilon$-\VSRL and \PLPG can do, instead of completely relying on sensors such as \VSRL. 
Hence, compared to the safety-agnostic baseline \PPO, while $\epsilon$-\VSRL and \PLPG respectively reduce violations by 37.8\% and 50.2\%, \VSRL causes 18\% more violations even when given perfect safety sensors.
Note that we measure the actual violations instead of policy safety $P_{\pi}(\mathtt{safe}|s)$. 
The policy safety during the learning process is plotted in \cref{app: experiment details}.

\subsection*{Q2: Selection of Hyperparameters $\alpha$ and $\epsilon$}
\begin{figure}[h]
    \centering
    \includegraphics[width=\columnwidth]{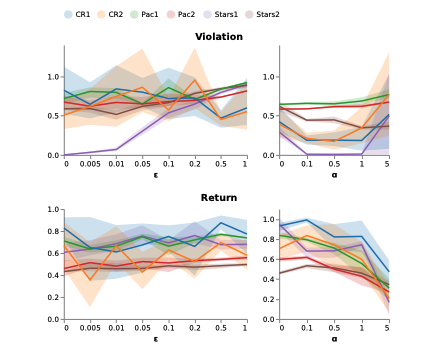}
    \caption{The episodic return and cumulative constraint violation of \VSRL (left) and \PLPG (right) agents in noisy environments. \textbf{Left}: The effect of $\epsilon$ is not significant. \textbf{Right}: Increasing $\alpha$ gives a convex trend in both return and violation. The absolute numbers are listed in \cref{app: experiment details}. }\label{fig: hyperparameter analysis}
\end{figure}

We analyze how the hyper-parameters $\alpha$ and $\epsilon$ respectively affect the performance of \PLPG and \VSRL in noisy environments. 
Specifically, we measure the episodic return and cumulative constraint violation. The results are plotted in \cref{fig: hyperparameter analysis}. 

The effect of different values of $\alpha$ to \PLPG agents is clear. 
Increasing $\alpha$ gives a convex trend in both the return and the violation counts. 
The optimal value of $\alpha$ is generally between $0.1$ and $1$, where the cumulative constraint violation and the return are optimal.
This illustrates the benefit of combining the shielded policy gradient and the safety gradient.  
On the contrary, the effect of $\epsilon$ to \VSRL agents is not significant. Increasing $\epsilon$ generally improves return but worsens constraint violation. This illustrates that simply randomizing the policy is not effective in improving the ability of handling noisy environments. Notice that there is no one-to-one mapping between the two hyper-parameters, as $\alpha$ controls the combination of the gradients and $\epsilon$ controls the degree of unsafe 
exploration.

\subsection*{Q3: PLPG Gradient Analysis}

\begin{figure}[ht]
    \centering
    \includegraphics[width=\columnwidth]{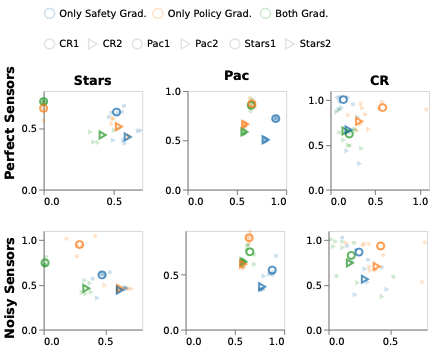}
    \caption{Trade-off between \textbf{Violation (x-axis) and Return (y-axis)}. Each data point is an agent's policy and the big data points are the average of five seeds.  }\label{fig: return-violation-LTST}
\end{figure}

We analyze how the shielded policy gradient and the safety gradient interact with one another. 
To do this, we introduce two ablated agents, one uses only safety gradients (\cref{eq: shielded policy gradient}) and the other uses only policy gradients (\cref{eq: safety gradient}). The results are plotted in \cref{fig: return-violation-LTST} where each data point is a policy trained for 600k steps. 

\cref{fig: return-violation-LTST} illustrates that combining both gradients is safer than using only one gradient. 
When augmented with perfect sensors, the use of both gradients has a violation that is lower than using only safety or policy gradient by 24.7\% and 10.8\%, respectively. 
When augmented with noisy sensors, using both gradients has a violation that is lower than using only safety or policy gradient by 25.5\% and 15.5\%, respectively. 
Nonetheless, the importance of policy and safety gradients varies among domains. In Stars and Pacman, using only policy gradients causes fewer violations than using only safety gradients. In CR, however, using only safety gradients causes fewer violations. This is a consequence of the inertia of the car not being fully captured by safety sensors, as discussed in Q1. 

\subsection*{Q4: Multi-step Safety Look-ahead}
We analyze the behaviour of PLPG when we use probabilistic logic shields for multi-step safety look-ahead. 
In this case, the safety program requires more sensors around the agent for potential danger over a larger horizon in the future.
In Pacman, for example, new sensors
are required to detect whether there is a ghost N units away where N is the safety horizon. We analyze the behavior/performance in terms of return/violation and computational cost. The results are shown in \cref{tab: Q4}. 
Increasing the horizon causes an exponential growth in the compiled circuit size and the corresponding inference time. However, we can considerably improve both safety and return, especially when moving from one to two steps.

\begin{table}[]\small
\begin{tabular}{@{}ccccc@{}}
\toprule
Safety horizon & 1 step & 2 steps & 3 steps & 4 steps \\ \midrule
\#sensors          & 4      & 12      & 24      & 40      \\ \midrule
Circuit size       & 116    & 1073    & 8246    & 373676  \\ \midrule
Compilation (s)   & 0.29   & 0.70    & 1.70    & 15.94   \\ \midrule
Evaluation (s)    & 0.01   & 0.08    & 0.66    & 27.39   \\ \midrule
Ret./Vio.  & 0.81 / 0.82  &  0.85 / 0.65   &  \textbf{0.86 / 0.57}   &  -   \\ \bottomrule
\end{tabular}
\caption{Multi-step Safety Look-ahead }\label{tab: Q4}
\end{table}

\section{Related Work}
\paragraph{Safe RL.} 
Safe RL aims to avoid unsafe consequences through the use of various safety representations ~\cite{Garcia2015}. 
There are several ways to achieve this goal.
One could constrain the expected cost~\cite{Achiam2017,Moldovan2012}, maximize safety constraint satisfiability through a loss function~\cite{SemanticLoss2018},
add a penalty to the agent when the constraint is violated~\cite{Pham2017,Tessler2018,Memarian2021}, or
construct a more complex reward structure using temporal logic~\cite{DeGiacomo2019,Camacho2019,Jiang2020,Hasanbeig2019,DenHengst2022}. 
These approaches express safety as a loss, while our method directly prevents the agent from taking actions that can potentially lead to safety violation. 

\paragraph{Probabilistic Shields.} 
Shielding is a safe reinforcement learning approach that aims to completely avoid unsafe actions during the learning process~\cite{DetShield2017,Shielding2020}. 
Previous shielding approaches have been limited to symbolic state spaces and are not suitable for noisy environments~\cite{Shielding2020,Harris2020,HardNeuralShielding2020,Anderson2020}. 
To address uncertainty, some methods incorporate randomization, e.g. simulating future states in an emulator to estimate risk~\cite{Li2020,Giacobbe2021}, using $\epsilon$-greedy exploration that permits unsafe actions~\cite{Garcia2019}, or randomizing the policy based on the current belief state~\cite{Karkus2017}. 
To integrate shielding with neural policies, one can translate a neural policy to a symbolic one that can be formally verified~\cite{Bastani2018,Verma2019}. 
However, these methods rely on sampling and do not have a clear connection to uncertainty present in the environment while our method directly exploits such uncertainty through the use of PLP principles. 
Belief states can capture uncertainty but require an environment model to 
to keep track of the agent's belief state, which is a stronger assumption than our method~\cite{Junges2021,Carr2022}.

\paragraph{Differentiable layers for neural policies.}
The use of differentiable shields has gain some attention in the field. 
One popular approach is to add a differentiable layer to the policy network to prevent constraint violations. 
Most of these methods focus on smooth physical rules~\cite{Dalal2018,Pham2017,Cheng2019} and only a few involve logical constraints.
In \cite{Kimura2021b}, an optimization layer is used to transform a state-value function into a policy encoded as a logical neural network. 
\cite{Ahmed2022} encode differentiable and hard logical constraints for neural networks using PLP. While being similar, this last model focuses on prediction tasks and do not consider a trade-off between return and constraint satisfiability.

\section{Conclusion}
\label{sec:conclusions}
We introduced Probabilistic Logic Shields, a proof of concept of a novel class of end-to-end differentiable shielding techniques. 
PLPG enable efficient training of safe-by-construction neural policies. 
This is done by using a probabilistic logic programming layer on top of the standard neural policy.
PLPG is a generalization of classical shielding \cite{HardNeuralShielding2020} that allows for both deterministic and probabilistic safety specifications. 
Future work will be dedicated to extending PLS to be used with a larger class of RL algorithms such as the ones with continuous policies.

\bibliographystyle{named}
\bibliography{main}

\newpage

\newpage

\appendix
\section{Experiment settings}
\label{app: experiments}
Experiments are run on machines that consist of Intel(R) Xeon(R) E3-1225 CPU cores and 32Gb memory. 
All environments in this work are available on GitHub under the MIT license.
The measurements are obtained by taking the average of five seeds.  
All agents are trained using PPO in \texttt{stable-baselines}~\cite{stable-baselines} with batch size=512, n\_epochs=15, n\_steps=2048, clip range=0.1, learning rate=0.0001. All policy networks and value networks have two hidden layers of size 64. 
All the other hyperparameters are set to default as in \texttt{stable-baselines}~\cite{stable-baselines}. 
The source code will be available upon publication of the paper.

\subsection{Stars} 

\begin{figure}[ht]
\centering
\begin{subfigure}[b]{0.3\columnwidth}
    \centering
    \includegraphics[width=\columnwidth]{Stars.png}
\end{subfigure}\hfill
\begin{subfigure}[b]{0.3\columnwidth}
    \centering
    \includegraphics[width=\columnwidth]{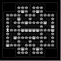}
\end{subfigure}\hfill
\begin{subfigure}[b]{0.3\columnwidth}
    \centering
    \includegraphics[width=\columnwidth]{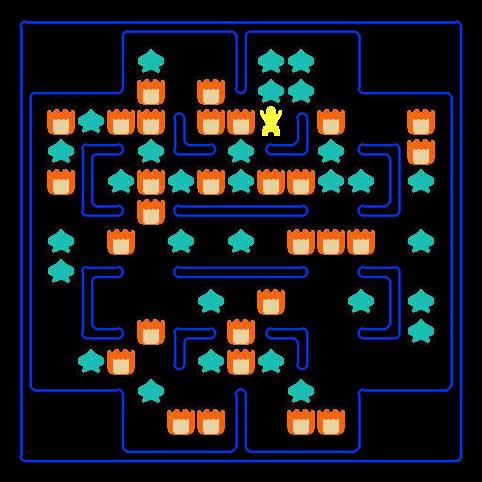}
\end{subfigure}
\caption{\textbf{Left}: A Stars image. \textbf{Middle}: A Stars state (downsampled from the left). \textbf{Right}: A example to pre-train noisy sensors. }
\label{fig: Stars state}
\end{figure}
We build the Stars environments using Berkeley's Pac-Man Project \footnote{\url{http://ai.berkeley.edu/project_overview.html}}. Our environment is a 15 by 15 grid world containing stars and fires, e.g. \cref{fig: Stars state}, where the agent can move around using five discrete, deterministic actions: stay, up, down, left, right. Each action costs a negative reward of $-0.1$. 
The task of the agent is to collect all stars in the environment. 
Collecting one star yields a reward of $1$.
When the agent finds all stars, the episode ends with a reward of $10$.
If the agent crashes into fire or has taken $200$ steps (maximum episode length), the episode ends with no rewards.
Each state is a downsampled, grayscale image where each pixel is a value between -1 and 1, e.g. \cref{fig: Stars state} (right). 

We use the following probabilistic logic shield. 
The $\mathtt{act/1}$ predicates represent the base policy and the $\mathtt{fire/2}$ predicates represent whether there is fire in an immediate neighboring grid.
These predicates are neural predicates, meaning that all probabilities $\mathtt{a_i}$ (resp. $\mathtt{f_i}$ will be substituted by real values in $[0, 1]$, produced by the underlying base policy (resp. perfect or noisy sensors). 
In \cref{fig: Stars state} (right), the sensors may produce $\{\mathtt{0::fire(0, 1)}$, $\mathtt{0::fire(0, -1)}$, $\mathtt{1::fire(-1, 0)}$, $\mathtt{0::fire(1, 0)}\}$.

\begin{verbatim}
% BASE POLICY
a0::act(stay); a1::act(up);   
a2::act(down); a3::act(left); 
a4::act(right).

% SENSORS
f0::fire( 0, 1).  f1::fire( 0,-1). 
f2::fire(-1, 0).  f3::fire( 1, 0).

% SAFETY LOOKAHEAD 
xagent(stay, 0, 0).  xagent(left,-1, 0).  
xagent(right,1, 0).  xagent(up,   0, 1).  
xagent(down, 0,-1).

% SAFETY CONSTRAINT
crash:- act(A), 
        xagent(A, X, Y), fire(X, Y).
safe:- not(crash).
\end{verbatim}

\subsection{Pacman}
Pacman is more complicated than the Stars in that the fire ghosts can move around and their transition models are unknown. All the other environmental parameters are the same as Stars.

We use the following probabilistic logic shield to perform a two-step look-ahead. The agent has 12 sensors to detect whether there is a fire ghost one or two units away. The agent is assumed to follow the policy at $T=0$ and then do nothing at $T=1$. The fire ghosts are assumed to be able to move to any immediate neighboring cells at $T=0$ and $T=1$. A crash can occur at $T=0$ or $T=1$.
\begin{verbatim}
% BASE POLICY
a0::act(stay); a1::act(up);   
a2::act(down); a3::act(left); 
a4::act(right).

% SENSORS 
f0:: fire(0, 0, 1). f1:: fire(0, 0,-1).
f2:: fire(0,-1, 0). f3:: fire(0, 1, 0).
f4:: fire(0, 0, 2). f5:: fire(0, 0,-2).
f6:: fire(0,-2, 0). f7:: fire(0, 2, 0).
f8:: fire(0, 1, 1). f9:: fire(0, 1,-1).
f10::fire(0,-1, 1). f11::fire(0,-1,-1).

% SAFETY LOOKAHEAD
fire(T, X, Y):-
    fire(T-1, PrevX, PrevY),
    move(PrevX, PrevY, _, X, Y).

agent(1, X, Y):-
    action(A), move(0, 0, A, X, Y).

agent(2, X, Y):- agent(1, X, Y).

move(X, Y, stay,  X,   Y).
move(X, Y, left,  X-1, Y).
move(X, Y, right, X+1, Y).
move(X, Y, up,    X,   Y+1).
move(X, Y, down,  X,   Y-1).

% SAFETY CONSTRAINT
crash:- fire(T, X, Y), agent(T, X, Y).
safe:- not(crash).
\end{verbatim}

\subsection{Car Racing} 
\begin{figure}[ht]
\centering
\begin{subfigure}[b]{0.33\columnwidth}
    \centering
    \includegraphics[width=\columnwidth]{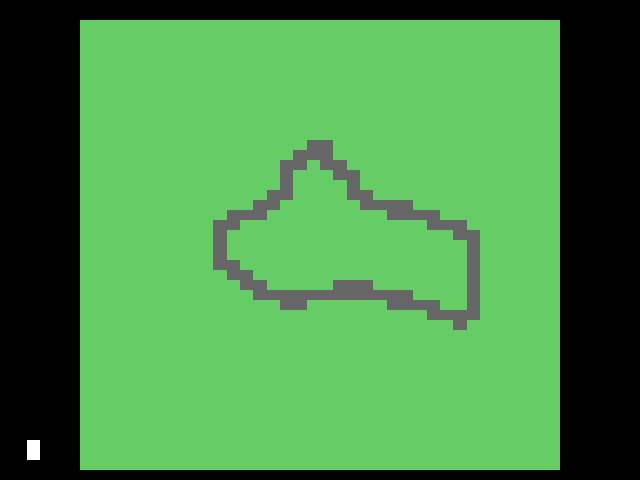}
\end{subfigure}\hfill
\begin{subfigure}[b]{0.33\columnwidth}
    \centering
    \includegraphics[width=\columnwidth]{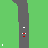}
\end{subfigure}\hfill
\begin{subfigure}[b]{0.33\columnwidth}
    \centering
    \includegraphics[width=\columnwidth]{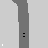}
\end{subfigure}
\caption{\textbf{Left}: A lap in Car Racing (Configuration 1). \textbf{Middle}: A CR image. We mark the sensors around the agent white but they not visible in training. \textbf{Right}: A CR state (downsampled from the middle). }
\label{fig: CR state}
\end{figure}

We simplify Gym's Car Racing environment~\cite{Brockman2016openai}. Our environment is a randomly generated car track, e.g. \cref{fig: CR state} (left).
At the beginning, the agent will be put on the track, e.g. \cref{fig: CR state} (middle).
The task of the agent is to drive around the track (i.e. to finish one lap) within 1000 frames. 
In each frame, the agent takes a discrete action: do-nothing, accelerate, brake, turn-left or turn-right. 
Each action costs a negative reward of $-0.1$. 
The agent gets a small reward if it continues to follow the track.
The agent does not get a penalty for driving on the grass, however, if the car drives outside of the map (i.e. the black part in \cref{fig: CR state}, left), the episode ends with a negative reward of $-100$. 
Each state is a downsampled, grayscale image where each pixel is a value between $-1$ and $1$, e.g. \cref{fig: CR state} (right). 

The original environment has a continuous action space that consists of three parameters: steering, gas and break with the minimum values $[-1, 0, 0]$ and the maximum values $[+1, +1, +1]$. In this paper, all agents use five discrete actions: do-nothing ($[0, 0, 0]$), accelerate ($[0, 1, 0]$), brake ($[0, 0, 0.8]$), turn-left ($[-1, 0, 0]$), turn-right ($[1, 0, 0]$). 

We use the following probabilistic logic shield.
The $\mathtt{act/1}$ predicates represent the base policy and the $\mathtt{grass/1}$ predicates represent whether there is grass in front, on the left or right hand side of the agent. 
In \cref{fig: CR state} (middle), the sensors may produce $\{\mathtt{0::grass(0)}$, $\mathtt{0::grass(1)}$ $\mathtt{0::grass(2)}\}$.


\begin{verbatim}
% BASE POLICY
a0::act(dn);   a1::act(accel); 
a2::act(brake);
a3::act(left); a4::act(right).

% SENSORS
g0::grass(front).  g1::grass(left).
g2::grass(right).

% SAFETY LOOKAHEAD 
ingrass:- 
    grass(left), not(grass(right)), 
    (act(left); act(accel)).
ingrass:- 
    not(grass(left)), grass(right), 
    (act(right); act(accel)).

% SAFETY CONSTRAINT
safe:- not(ingrass).
\end{verbatim}

\subsection{Approximating Noisy Sensors}
We approximate noisy sensors using convolutional neural networks.  In all the environments, we use 4 convolutional layers with respectively 8,16,32,64 (5x5) filters and relu activations. The output is computed by a dense layer with sigmoid activation function. The number of output neurons depends on the ProbLog program for that experiment (see Appendix A.1).
We pre-train the networks and then we fix them during the reinforcement learning process. 
The pre-training strategy is the following. We generated randomly 3k images for Stars and Pacman with 30 fires and 30 stars, e.g. \cref{fig: Stars state} (right), and 2k images for Car Racing, e.g. \cref{fig: CR state} (right). We selected the size of the pre-training datasets in order to achieve an accuracy higher than 99\% on a validation set of 100 examples. 


\section{Safety Guarantees of Probabilistic Logic Shield}
\label{app: proof}
In \cref{def: differentiable shield}, a shielded policy $\pi^+$ is always safer than its base policy $\pi$ for all $s$ and $\pi$, i.e.
\begin{align*}
    \mathbf{P}_{\pi^+}(\mathtt{safe}|s) \geq \mathbf{P}_{\pi}(\mathtt{safe}|s)\mbox{      }
\forall \pi\in \Pi\mbox{ and } \forall s\in S
\end{align*}

\begin{proof}
\begin{align*}
    &\mathbf{P}_{\pi^+}(\mathtt{safe}|s) 
    \tag*{$\triangleright$ \cref{eq: policy safety}} \\
= &\sum_{a\in A} \pi^+(a|s)\mathbf{P}(\mathtt{safe}|s, a) 
\tag*{$\triangleright$ \cref{eq: shielded policyy}}\\
= &\frac{1}{\mathbf{P}_{\pi}(\mathtt{safe}|s)}[\sum_{a\in A} \pi(a|s)\mathbf{P}(\mathtt{safe}|s, a)^2]
\tag*{$\triangleright$ Jensen's inequality}\\
\geq &\frac{1}{\mathbf{P}_{\pi}(\mathtt{safe}|s)}[\sum_{a\in A} \pi(a|s)\mathbf{P}(\mathtt{safe}|s, a)]^2 
\tag*{$\triangleright$ \cref{eq: policy safety}}\\
= &\mathbf{P}_{\pi}(\mathtt{safe}|s) 
\end{align*}
The inequality comes from Jensen's inequality \cite{pishro2014introduction}, which states that $\mathbb{E}[g(X)]\geq g(\mathbb{E}[X])$ for any convex function $g(X)$. In this proof, $g(X)= X^2$.
\end{proof}

\section{A Compiled Circuit}
\label{app: circuit}
\begin{figure}[h]
    \centering
    \includegraphics[width=0.9\columnwidth]{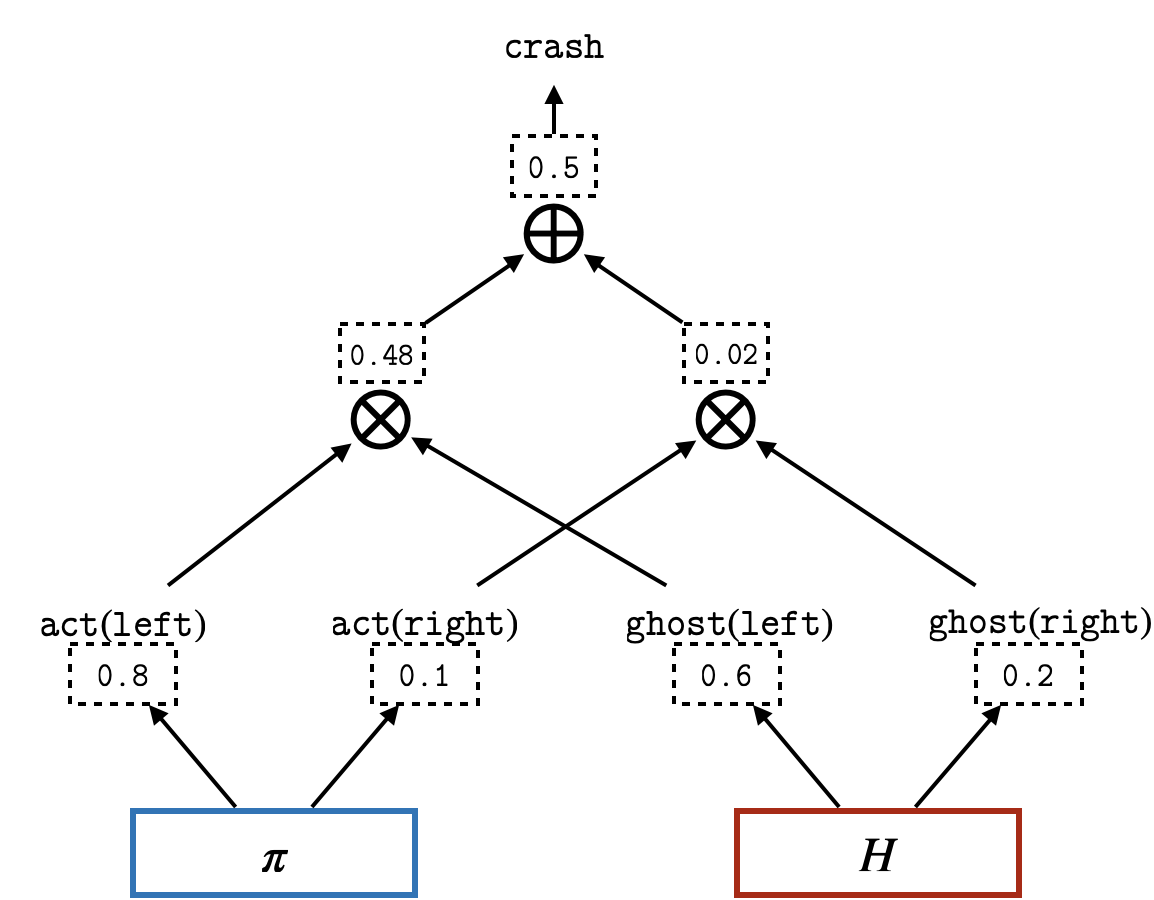}
    \caption{A circuit compiled from a program.}
    \label{fig: circuit}
\end{figure}

The following program is compiled to the circuit in \cref{fig: circuit}.
\begin{verbatim}
0.2::act(dn); 0.6::act(left); 
0.2::act(right).
0.8::ghost(left).  0.1::ghost(right).
crash:- act(left), ghost(left).
crash:- act(right), ghost(right).
\end{verbatim}

\begin{table*}[]
\centering
\begin{tabular}{@{}ccccccc@{}}
\toprule
\multirow{2}{*}{} & \multirow{2}{*}{Description} & \multirow{2}{*}{Return Range} & \multirow{2}{*}{Violation Range} & Perfect Sensors & \multicolumn{2}{c}{Noisy Sensors} \\
                  &                              &                               &                                  & $\alpha$        & $\epsilon$       & $\alpha$       \\ \midrule
Stars1            & Deterministic actions        & \multirow{2}{*}{{[}0, 45{]}}  & \multirow{2}{*}{{[}0, 15k{]}}    & 0.5             & 0.005            & 1              \\ \cmidrule(r){1-2} \cmidrule(l){5-7} 
Stars2            & Stochastic actions           &                               &                                  & 1               & 0.01             & 1              \\ \midrule
Pacman1           & One ghost                    & \multirow{2}{*}{{[}0, 40{]}}  & {[}0, 7k{]}                      & 0.1             & 0.05             & 0.5            \\ \cmidrule(r){1-2} \cmidrule(l){5-7} 
Pacman2           & Two ghosts                   &                               & {[}0, 15k{]}                     & 0.5             & 0.005            & 0.1            \\ \midrule
Car Racing1       & Smoother track               & \multirow{2}{*}{{[}0, 900{]}} & \multirow{2}{*}{{[}0, 1{]}}      & 0.5             & 0.5              & 1              \\ \cmidrule(r){1-2} \cmidrule(l){5-7} 
Car Racing2       & Curlier track                &                               &                                  & 0.1             & 0.5              & 0.5            \\ \bottomrule
\end{tabular}
\caption{Summary of configurations. }
\label{tab: configurations}
\end{table*}

\newpage
\section{Experiment Details}
\label{app: experiment details}

We run two configurations for each domain. The second configuration is more challenging than the first one. The configurations may have different normalization ranges and hyperparameters. We select the hyperparameters that result in the lowest violations in each configuration. 
\cref{tab: configurations} lists a summary of all configurations and all other tables follows this table if not explicitly specified.

We train all agents in all configurations using 600k learning steps and all experiments are repeated using five different seeds. The following tables list the normalized, average return and violation values of the five runs.
\cref{tab: main} lists the return/violation results of all agents (cf. the large data points in \cref{fig: return-violation}).  
\cref{tab: return-violation-ablation} lists the return/violation results for \PLPG gradient analysis  (cf. the large data points in \cref{fig: return-violation-LTST}).
\cref{tab: alpha choices noisy,tab: eps choices} respectively list the return/violation results for $\alpha\in\{0, 0.1, 0.5, 1, 5\}$ and $\epsilon\in\{0, 0.005, 0.01, 0.05, 0.1, 0.2, 0.5, 1.0\}$ (cf. \cref{fig: hyperparameter analysis}).

\begin{table*}[]
\centering
\begin{tabular}{@{}ccccccc@{}}
\toprule
\multirow{2}{*}{Return/Violation} & No Safety                & \multicolumn{2}{c}{Perfect Sensors}               & \multicolumn{3}{c}{Noisy Sensors}                                   \\ \cmidrule(l){2-7} 
                                  & \multicolumn{1}{c|}{PPO} & VSRL                 & \multicolumn{1}{c|}{PLPG} & VSRL                 & $\epsilon$-VSRL      & PLPG                 \\ \midrule
Stars1                            & 0.68 / 0.90              & 0.57 / \textbf{0.00} & \textbf{0.71 / 0.00}      & 0.61 / \textbf{0.00} & 0.64 / 0.03          & \textbf{0.74} / 0.01 \\ \midrule
Stars2                            & 0.50 / 0.89              & 0.43 / 0.58          & \textbf{0.44 / 0.42}      & 0.43 / 0.59          & 0.46 / 0.52          & \textbf{0.46 / 0.35} \\ \midrule
Pacman1                           & 0.74 / 0.92              & 0.72 / 0.74          & \textbf{0.85 / 0.65}      & 0.71 / 0.72          & \textbf{0.75 / 0.65} & 0.71 / \textbf{0.65} \\ \midrule
Pacman2                           & 0.56 / 0.82              & 0.56 / 0.62          & \textbf{0.59 / 0.58}      & 0.46 / 0.67          & 0.51 / 0.62          & \textbf{0.62 / 0.59} \\ \midrule
Car Racing1                       & 0.77 / 0.60              & \textbf{0.74} / 0.84 & 0.62 / \textbf{0.21}      & 0.83 / 0.82          & \textbf{0.88} / 0.47 & 0.83 / \textbf{0.18} \\ \midrule
Car Racing2                       & 0.58 / 0.55              & 0.55 / 0.67          & \textbf{0.65 / 0.17}      & 0.67 / 0.51          & 0.70 / 0.45          & \textbf{0.74 / 0.17} \\ \bottomrule
\end{tabular}
\caption{Normalized return and violation for all agents (cf. the large data points in \cref{fig: return-violation}). 
}\label{tab: main}
\end{table*}

\begin{table*}[]
\begin{tabular}{@{}ccccccc@{}}
\toprule
\multirow{2}{*}{Return/Violation} & \multicolumn{3}{c}{Perfect Sensors}                                     & \multicolumn{3}{c}{Noisy Sensors}                              \\ \cmidrule(l){2-7} 
                                  & Only Safety Grad     & Only Policy Grad     & \multicolumn{1}{c|}{Both} & Only Safety Grad & Only Policy Grad     & Both                 \\ \midrule
Stars1                            & 0.63 / 0.52          & 0.66 / \textbf{0.00} & \textbf{0.71 / 0.00}      & 0.61 / 0.47      & \textbf{0.95} / 0.29 & 0.74 / \textbf{0.01} \\ \midrule
Stars2                            & 0.43 / 0.60          & \textbf{0.51} / 0.53 & 0.44 / \textbf{0.42}      & 0.44 / 0.62      & \textbf{0.46} / 0.62 & \textbf{0.46 / 0.35} \\ \midrule
Pacman1                           & 0.72 / 0.89          & \textbf{0.87 / 0.65} & 0.85 / \textbf{0.65}      & 0.54 / 0.88      & \textbf{0.84 / 0.64} & 0.71 / 0.65          \\ \midrule
Pacman2                           & 0.51 / 0.79          & \textbf{0.66 / 0.58} & 0.59 / \textbf{0.58}      & 0.39 / 0.78      & \textbf{0.60 / 0.58} & \textbf{0.62} / 0.59 \\ \midrule
Car Racing1                       & \textbf{1.00 / 0.14} & 0.91 / 0.58          & 0.62 / 0.21               & 0.86 / 0.25      & \textbf{0.93} / 0.42 & 0.83 / \textbf{0.18} \\ \midrule
Car Racing2                       & 0.67 / 0.20          & \textbf{0.76} / 0.32 & 0.65 / \textbf{0.17}      & 0.56 / 0.29      & 0.70 / 0.39          & \textbf{0.74 / 0.17} \\ \bottomrule
\end{tabular}
\caption{Analysis of PLPG gradients (cf. the large data points in \cref{fig: return-violation-LTST}). }\label{tab: return-violation-ablation}
\end{table*}

\begin{table*}[]
\centering
\begin{tabular}{@{}cccccc@{}}
\toprule
$\alpha$    & 0                    & 0.1         & 0.5                  & 1                    & 5           \\ \midrule
Stars1      & 0.95 / 0.29          & 0.68 / 0.01 & 0.68 / 0.01          & \textbf{0.74 / 0.01} & 0.17 / 0.50 \\ \midrule
Stars2      & 0.46 / 0.62          & 0.53 / 0.44 & 0.51 / 0.45          & \textbf{0.46 / 0.35} & 0.34 / 0.37 \\ \midrule
Pacman1     & 0.84 / 0.64          & 0.79 / 0.66 & \textbf{0.71 / 0.65} & 0.56 / 0.69          & 0.31 / 0.77 \\ \midrule
Pacman2     & 0.60 / 0.58 & \textbf{0.62 / 0.59} & 0.50 / 0.62          & 0.43 / 0.62          & 0.27 / 0.67 \\ \midrule
Car Racing1 & 0.93 / 0.42          & 0.99 / 0.19 & 0.82 / 0.19          & \textbf{0.83 / 0.18} & 0.47 / 0.52 \\ \midrule
Car Racing2 & 0.70 / 0.39          & 0.84 / 0.21 & \textbf{0.74 / 0.17} & 0.60 / 0.34          & 0.21 / 0.77 \\ \bottomrule
\end{tabular}
\caption{The return and violation of different $\alpha$ values in noisy environments (cf. \cref{fig: hyperparameter analysis}). }\label{tab: alpha choices noisy}
\end{table*}

\begin{table*}[]
\centering
\begin{tabular}{@{}ccccccccc@{}}
\toprule
$\epsilon$  & 0 (VSRL)    & 0.005                & 0.01                 & 0.05                 & 0.1         & 0.2         & 0.5                  & 1 (PPO)     \\ \midrule
Stars1      & 0.61 / 0.00 & \textbf{0.64 / 0.03} & 0.69 / 0.07          & 0.76 / 0.30          & 0.69 / 0.54 & 0.76 / 0.65 & 0.68 / 0.80          & 0.68 / 0.90 \\ \midrule
Stars2      & 0.43 / 0.59 & 0.46 / 0.59          & \textbf{0.46 / 0.52} & 0.46 / 0.62          & 0.48 / 0.67 & 0.47 / 0.78 & 0.49 / 0.85          & 0.50 / 0.89 \\ \midrule
Pacman1     & 0.71 / 0.72 & 0.64 / 0.81          & 0.66 / 0.80          & \textbf{0.75 / 0.65} & 0.66 / 0.86 & 0.72 / 0.71 & 0.77 / 0.84          & 0.74 / 0.92 \\ \midrule
Pacman2     & 0.46 / 0.67 & \textbf{0.51 / 0.62} & 0.48 / 0.67          & 0.52 / 0.65          & 0.51 / 0.68 & 0.53 / 0.70 & 0.54 / 0.74          & 0.56 / 0.82 \\ \midrule
Car Racing1 & 0.83 / 0.82 & 0.65 / 0.65          & 0.61 / 0.84          & 0.67 / 0.80          & 0.75 / 0.72 & 0.66 / 0.73 & \textbf{0.88 / 0.47} & 0.77 / 0.60 \\ \midrule
Car Racing2 & 0.67 / 0.51 & 0.36 / 0.62          & 0.67 / 0.75          & 0.43 / 0.86          & 0.63 / 0.57 & 0.52 / 0.95 & \textbf{0.70 / 0.45} & 0.58 / 0.55 \\ \bottomrule
\end{tabular}
\caption{The return and violation of different $\epsilon$ values in noisy environments (cf. \cref{fig: hyperparameter analysis}). }\label{tab: eps choices}
\end{table*}

\begin{figure*}[htbp]
\centering
\caption{
Performance of agents augmented with perfect sensors. \textbf{Left:} The average episodic return. \textbf{Middle:} The cumulative episodic constraint violation. \textbf{Right:} The step-wise policy safety, i.e. $P_\pi(\mathtt{safe}|s)$. }\label{fig: perf sensors}
\includegraphics[width=\textwidth]{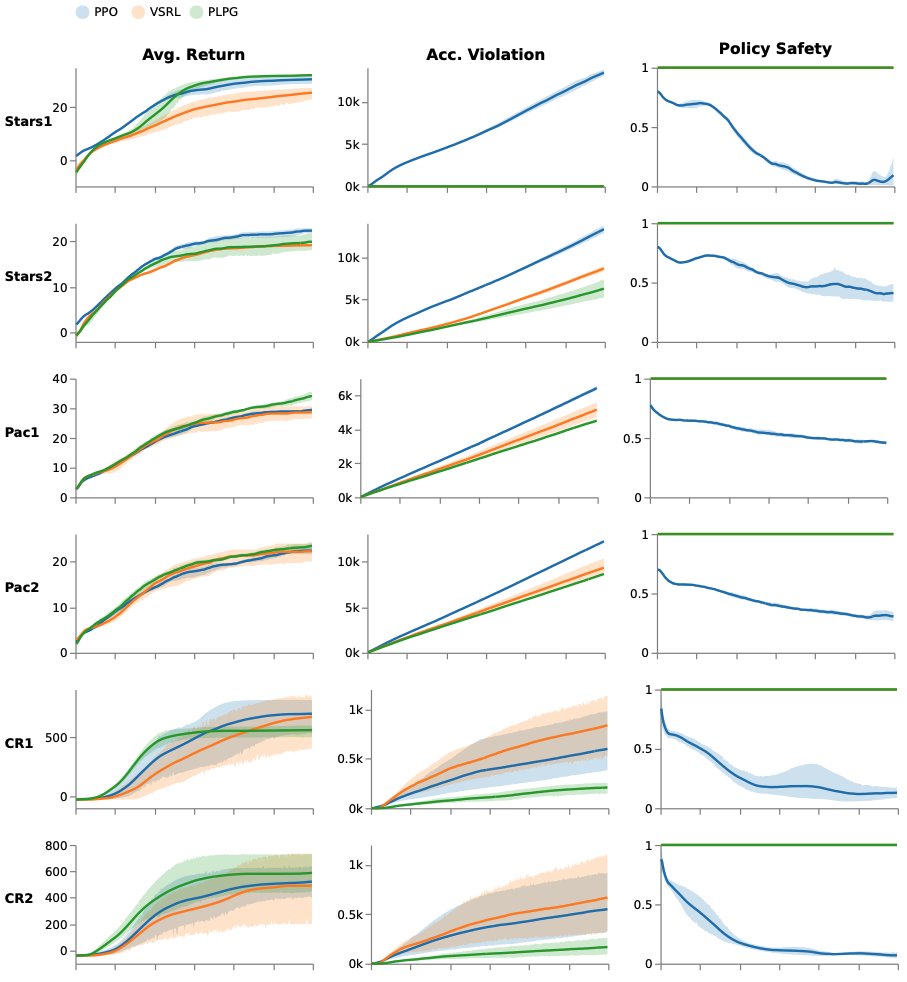}
\end{figure*}

\begin{figure*}[htbp]
\centering
\caption{
Performance of agents augmented with noisy sensors. \textbf{Left:} The average episodic return. \textbf{Middle:} The cumulative episodic constraint violation. \textbf{Right:} The step-wise policy safety, i.e. $P_\pi(\mathtt{safe}|s)$.
}\label{fig: noisy sensors}
\includegraphics[width=\textwidth]{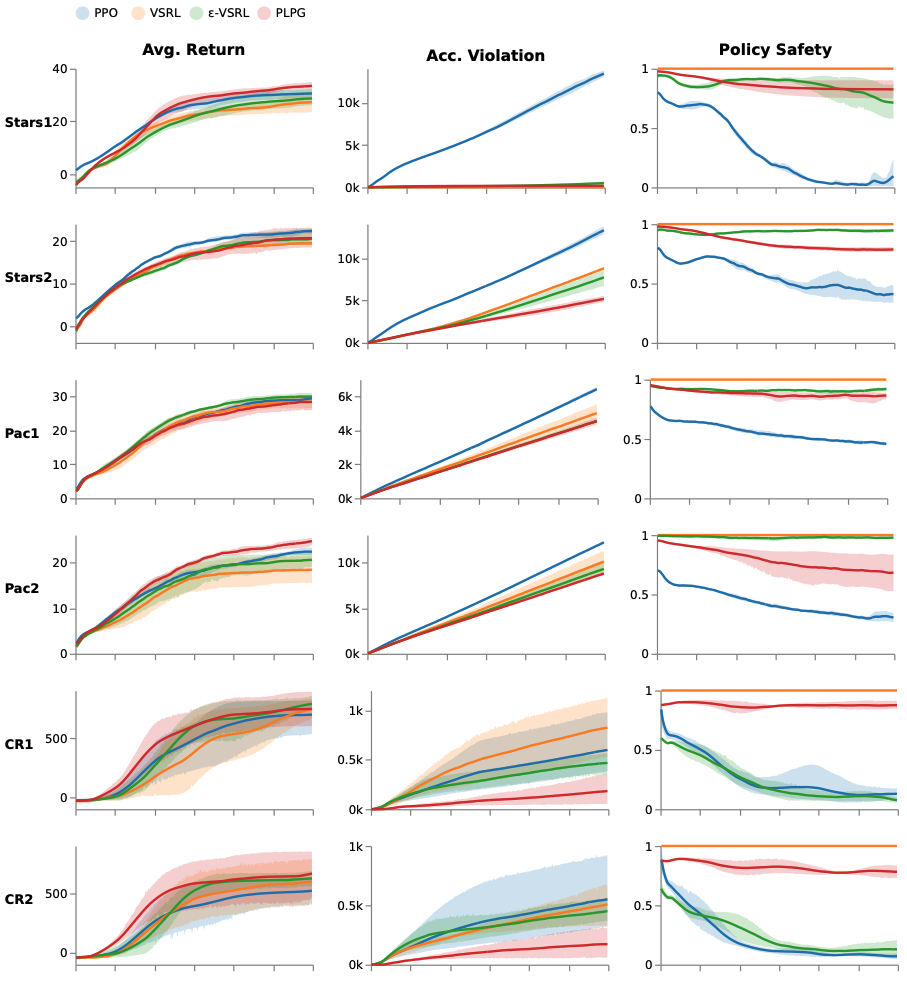}
\end{figure*}

We plot episodic return, cumulative violation and step-wise policy safety in the learning process for all agents ($\mathbf{P}_{\pi^+}(\mathtt{safe}|s)$ for shielded agents and $\mathbf{P}_{\pi}(\mathtt{safe}|s)$ for \PPO). \cref{fig: noisy sensors} plots these curves for the agents augmented with perfect sensors and \cref{fig: perf sensors} for the ones with noisy sensors. \PPO agents are plotted in both figures as a baseline but they are not augmented with any sensors. The return and policy safety curves are smoothed by taking the exponential moving average with the coefficient of 0.05.

\end{document}